\documentclass[letterpaper, 10 pt, conference]{ieeeconf}  % Comment this line out if you need a4paper

\IEEEoverridecommandlockouts                              % This command is only needed if 
\usepackage{graphicx} % for pdf, bitmapped graphics files
\usepackage{url} % assumes amsmath package installed
\usepackage{multirow}
\usepackage{hhline}
\title{\LARGE \bf
Real-Time Grasp Detection Using Convolutional Neural Networks
}

\author{Joseph Redmon$^{1}$, Anelia Angelova$^{2}$% <-this % stops a space
\thanks{$^{1}$University of Washington}%
\thanks{$^{2}$Google Research}%
}

\begin{document}

\maketitle
\thispagestyle{empty}
\pagestyle{empty}

%%%%%%%%%%%%%%%%%%%%%%%%%%%%%%%%%%%%%%%%%%%%%%%%%%%%%%%%%%%%%%%%%%%%%%%%%%%%%%%%
\begin{abstract}

We present an accurate, real-time approach to robotic grasp detection based on convolutional neural networks. Our network performs single-stage regression to graspable bounding boxes without using standard sliding window or region proposal techniques. The model outperforms state-of-the-art approaches by 14 percentage points and runs at 13 frames per second on a GPU. Our network can simultaneously perform classification so that in a single step it recognizes the object and finds a good grasp rectangle. A modification to this model predicts multiple grasps per object by using a locally constrained prediction mechanism. The locally constrained model performs significantly better, especially on objects that can be grasped in a variety of ways.
\end{abstract}

%%%%%%%%%%%%%%%%%%%%%%%%%%%%%%%%%%%%%%%%%%%%%%%%%%%%%%%%%%%%%%%%%%%%%%%%%%%%%%%%
\section{INTRODUCTION}

Perception---using the senses (or sensors if you are a robot) to understand your environment---is hard. Visual perception involves mapping pixel values and light information onto a model of the universe to infer your surroundings. General scene understanding requires complex visual tasks such as segmenting a scene into component parts, recognizing what those parts are, and disambiguating between visually similar objects. Due to these complexities, visual perception is a large bottleneck in real robotic systems.
   \begin{figure}[hbtp]
      \centering
        \includegraphics[width=\linewidth]{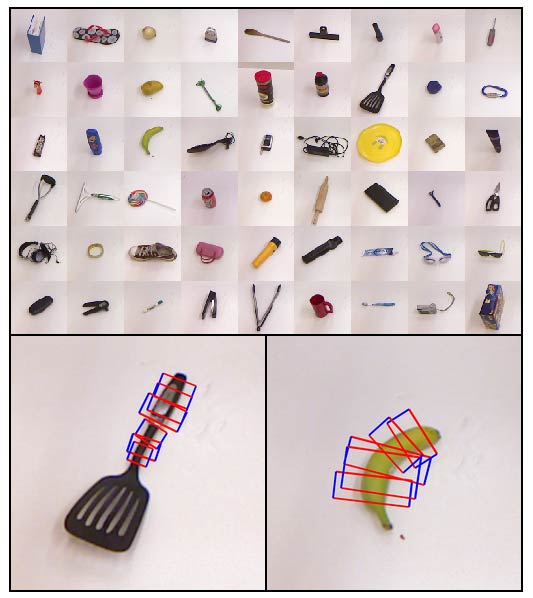}
      \caption{The Cornell Grasping Dataset contains a variety of objects, each with multiple labelled grasps. Grasps are given as oriented rectangles in 2-D.}
      \label{dataset}
   \end{figure}

General purpose robots need the ability to interact with and manipulate objects in the physical world. Humans see novel objects and know immediately, almost instinctively, how they would grab them to pick them up. Robotic grasp detection lags far behind human performance. We focus on the problem of finding a good grasp given an RGB-D view of the object.

We evaluate on the Cornell Grasp Detection Dataset, an extensive dataset with numerous objects and ground-truth labelled grasps (see Figure \ref{dataset}). Recent work on this dataset runs at 13.5 seconds per frame with an accuracy of 75 percent \cite{lenz2013deep} \cite{jiang2011efficient}. This translates to a 13.5 second delay between a robot viewing a scene and finding where to move its grasper.

The most common approach to grasp detection is a sliding window detection framework. The sliding window approach uses a classifier to determine whether small patches of an image constitute good grasps for an object in that image. This type of system requires applying the classifier to numerous places on the image. Patches that score highly are considered good potential grasps.

We take a different approach; we apply a single network once to an image and predict grasp coordinates directly. Our network is comparatively large but because we only apply it once to an image we get a massive performance boost. Instead of looking only at local patches our network uses global information in the image to inform its grasp predictions, making it significantly more accurate. Our network achieves 88 percent accuracy and runs at real-time speeds (13 frames per second). This redefines the state-of-the-art for RGB-D grasp detection.

\section{RELATED WORK}

Significant past work uses 3-D simulations to find good grasps \cite{bicchi2000robotic} \cite{miller2003automatic} \cite{miller2004graspit} \cite{pelossof2004svm} \cite{leon2010opengrasp}. These approaches are powerful but rely on a full 3-D model and other physical information about an object to find an appropriate grasp. Full object models are often not known a priori. General purpose robots may need to grasp novel objects without first building complex 3-D models of the object.

Robotic systems increasingly leverage RGB-D sensors and data for tasks like object recognition \cite{lai2011large}, detection \cite{lai2012detection} \cite{blum2012learned}, and mapping \cite{henry2010rgb} \cite{endres2012evaluation}. RGB-D sensors like the Kinect are cheap, and the extra depth information is invaluable for robots that interact with a 3-D environment.

Recent work on grasp detection focusses on the problem of finding grasps solely from RGB-D data \cite{saxena2008robotic}. These techniques rely on machine learning to find the features of a good grasp from data. Visual models of grasps generalize well to novel objects and only require a single view of the object, not a full physical model \cite{rao2010grasping} \cite{jiang2011efficient}.

Convolutional networks are a powerful model for learning feature extractors and visual models \cite{krizhevsky2012imagenet} \cite{girshick14CVPR}. Lenz et al. successfully use convolutional networks for grasp detection as a classifier in a sliding window detection pipeline \cite{lenz2013deep}. We address the same problem as Lenz et al. but use a different network architecture and processing pipeline that is capable of higher accuracy at much faster speeds.
   
\section{PROBLEM DESCRIPTION}

\begin{figure}[thpb]
      \centering
        \includegraphics[width=\linewidth]{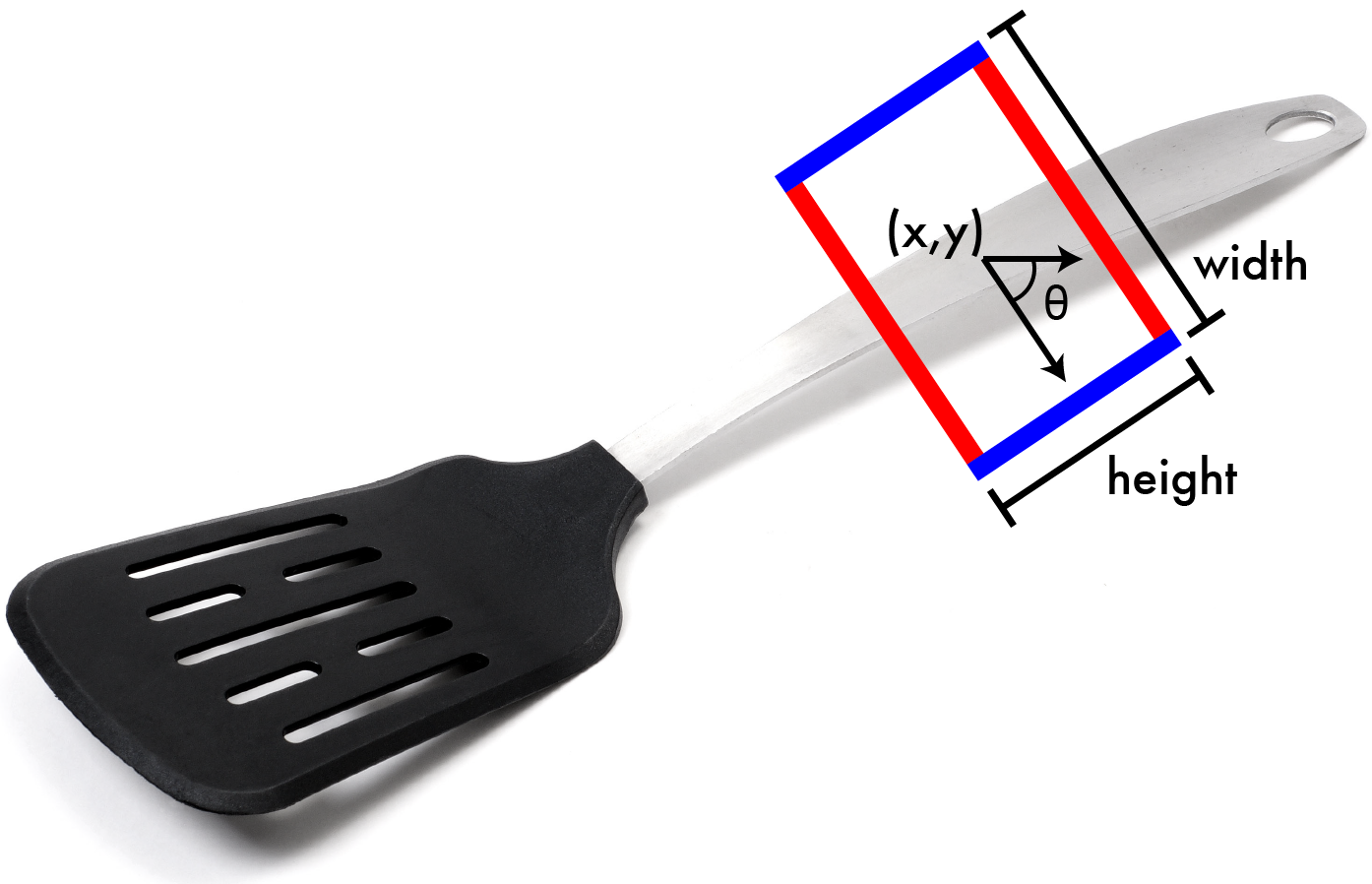}
      \caption{A five-dimensional grasp representation, with terms for location, size, and orientation. The blue lines demark the size and orientation of the gripper plates. The red lines show the approximate distance between the plates before the grasp is executed.}
      \label{grasp}
   \end{figure}

Given an image of an object we want to find a way to safely pick up and hold that object. We use the five-dimensional representation for robotic grasps proposed by Lenz et al. \cite{lenz2013deep}. This representation gives the location and orientation of a parallel plate gripper before it closes on an object. Ground truth grasps are rectangles with a position, size, and orientation:

\[g = \{x,y,\theta,h,w\}\]
where $(x,y)$ is the center of the rectangle, $\theta$ is the orientation of the rectangle relative to the horizontal axis, $h$ is the height, and $w$ is the width. Figure \ref{grasp} shows an example of this grasp representation.

This is a simplification of Jiang et al.'s seven-dimensional representation \cite{jiang2011efficient}. Instead of finding the full 3-D grasp location and orientation, we implicitly assume that a good 2-D grasp can be projected back to 3-D and executed by a robot viewing the scene. Lenz et al. describe a process to do this and while they don't evaluate it directly it appears to work well in their experiments \cite{lenz2013deep}.

Using a five-dimensional representation makes the problem of grasp detection analogous to object detection in computer vision with the only difference being an added term for gripper orientation.

\section{GRASP DETECTION WITH NEURAL NETWORKS}

   \begin{figure*}[thpb]
      \centering
        \includegraphics[width=\linewidth]{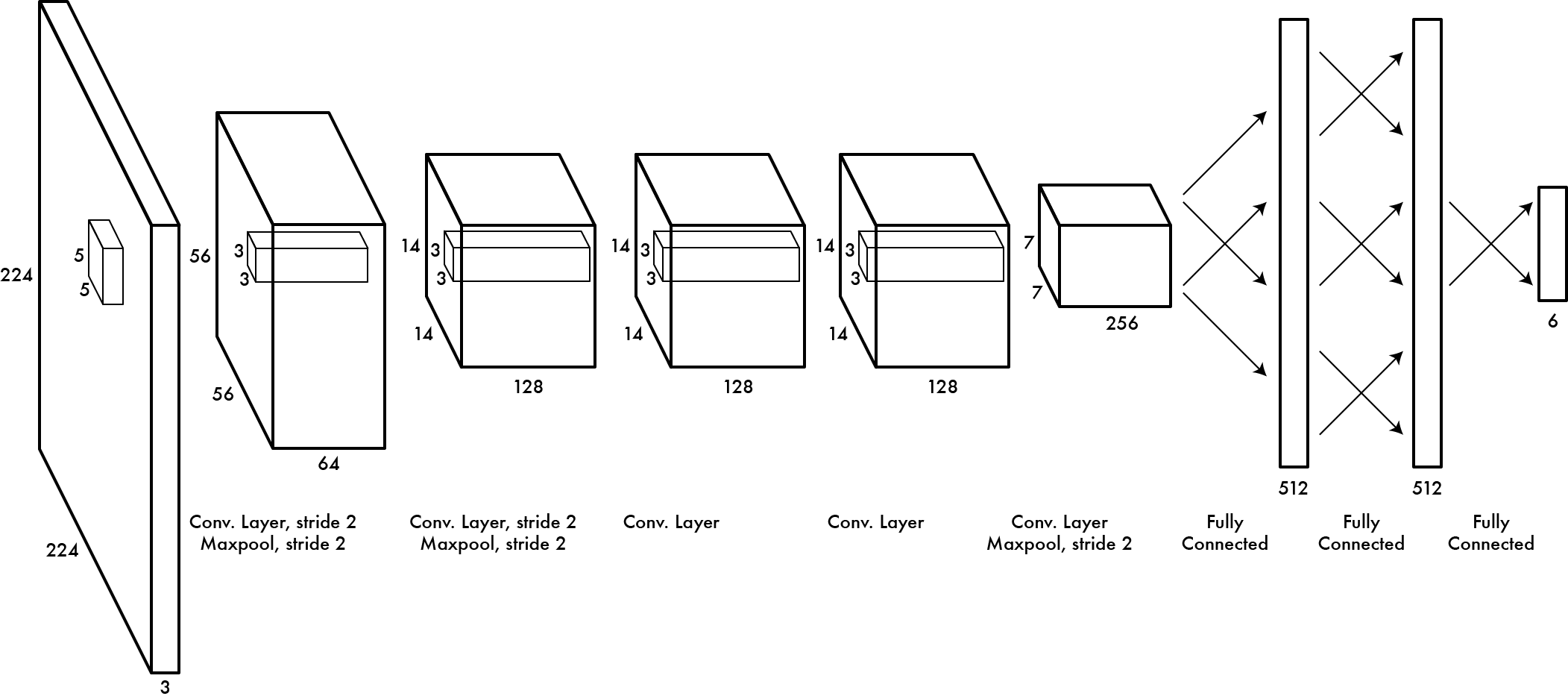}
      \caption{The full architecture of our direct regression grasp model.}
      \label{arch}
   \end{figure*}

Convolutional neural networks (CNNs) currently outperform other techniques by a large margin in computer vision problems such as classification \cite{krizhevsky2012imagenet} and detection \cite{girshick14CVPR}. CNNs already perform well on grasp detection when applied as a classifier in a sliding-window approach \cite{lenz2013deep}.

We want to avoid the computational costs of running a small classifier numerous times on small patches of an image. We harness the extensive capacity of a large convolutional network to make global grasp predictions on the full image of an object.

\subsection{Architecture}

When building our grasp detection system we want to start from a strong foundation. We derive our model from a version of the widely adopted convolutional network proposed by Krizhevsky et al. for object recognition tasks (AlexNet) \cite{krizhevsky2012imagenet}.

Our network has five convolutional layers followed by three fully connected layers. The convolutional layers are interspersed with normalization and maxpooling layers at various stages. A full description of the architecture can be found in Figure \ref{arch}.

\subsection{Direct Regression To Grasps}

The simplest model we explore is a direct regression from the raw RGB-D image to grasp coordinates. The raw image is given to the model which uses convolutional layers to extract features from the image. The fully connected layers terminate in an output layer with six output neurons corresponding to the coordinates of a grasp. Four of the neurons correspond to location and height. Grasp angles are two-fold rotationally symmetric so we parameterize by using the two additional coordinates: the sine and cosine of twice the angle.

This model assumes the strong prior that every image contains a single graspable object and it only needs to predict a one grasp for that object. This strong assumption may not hold outside of experimental conditions. In practice this model would have to come in a pipeline that first segments the image into pieces that only contain individual objects. The benefit of enforcing this assumption is that instead of classifying many of small patches in a sliding window type approach, we only need to look at a single image and make a global prediction.

During training our model picks a random ground truth grasp every time it sees an object to treat as the single ground truth grasp. Because the grasp changes often, the model does not overfit to a single grasp on an object. We minimize the squared error of the predicted grasp. The end effect is that our model fits to the average of the possible grasps for an object.

\subsection{Regression + Classification}

In order to use a grasped object the robot must first recognize the object. By extending our model we show that recognition and grasp detection can be combined into a single, efficient pipeline. 

We modify our architecture from the previous section by adding extra neurons to the output layer that correspond to object categories. We keep the rest of the architecture the same thus our model uses common features from the convolutional layers for both recognition and detection.

This combined model processes an image in a single pass and predicts both the category of the object in the image and a good grasp for that object. It runs just as fast as the direct regression model because the architecture remains largely unchanged.

\subsection{MultiGrasp Detection}

Our third model is a generalization of the first model, we call it MultiGrasp. The preceeding models assume that there is only a single correct grasp per image and try to predict that grasp. MultiGrasp divides the image into an NxN grid and assumes that there is at most one grasp per grid cell. It predicts one grasp per cell and also the likelihood that the predicted grasp would be feasible on the object. For a cell to predict a grasp the center of that grasp must fall within the cell.

The output of this model is an NxNx7 prediction. The first channel is a heatmap of how likely a region is to contain a correct grasp. The other six channels contain the predicted grasp coordinates for that region. For experiments on the Cornell dataset we used a 7x7 grid, making the actual output layer 7x7x7 or 343 neurons. Our first model can be seen as a specific case of this model with a grid size of 1x1 where the probability of the grasp existing in the single cell is implicitly one.

Training MultiGrasp requires some special considerations. Every time MultiGrasp sees an image it randomly picks up to five grasps to treat as ground truth. It constructs a heatmap with up to five cells marked with ones and the rest filled with zeros. It also calculates which cells those grasps fall into and fills in the appropriate columns of the ground truth with the grasp coordinates. During training we do not backpropagate error for the entire 7x7x7 grid because many of the column entries are blank (if there is no grasp in that cell). Instead we backpropagate error for the entire heatmap channel and also for the specific cells that contain ground truth grasps.

This model has several precursors in object detection literature but is novel in important aspects. Szegedy et al. use deep neural networks to predict binary object masks on images and use the predicted masks to generate bounding boxes \cite{szegedy2013detection}. The heatmap that we predict is similar to this object mask but we also predict full bounding boxes and only use the heatmap for weighting our predictions. Our system does not rely on post-processing or heuristics to extract bounding boxes but rather predicts them directly.

Erhan et al. predict multiple bounding boxes and confidence scores associated with those bounding boxes \cite{erhan2013scalable}. This approach is most similar to our own, we also predict multiple bounding boxes and weight them by a confidence score. The key difference is the we enforce structure on our predictions so that each cell can only make local predictions for its region of the image.

%\begin{table*}[tb]
%\caption{Simultaneous Detection and Classification Accuracy on the Cornell Dataset}
%\label{results}
%\begin{center}
%\begin{tabular}{|c|c|c|c|c|}
%\hline
%\multirow{2}{*}{\textbf{Algorithm}} & \multicolumn{2}{|c|}{\textbf{Detection}} & \multicolumn{2}{|c|}{\textbf{Classification}} \\
%\hhline{~----}
%& \textbf{Image-wise split} & \textbf{Object-wise split} & \textbf{Image-wise split} & \textbf{Object-wise split} \\
%\hline
%\hline
%Chance \cite{lenz2013deep} + Most Common Class & 6.7\% & 6.7\% & 17.7\% & 17.7\% \\
%\hline
%\textbf{Direct Regression} & 84.4\% & 84.9\% & - & -\\
%\hline
%\textbf{Regression + Classification} & 85.5\% & 84.9\% & 90.0\% & 61.5\% \\
%\hline
%\end{tabular}
%\end{center}
%\end{table*}

   \begin{figure*}[thpb]
      \centering
        \includegraphics[width=\linewidth]{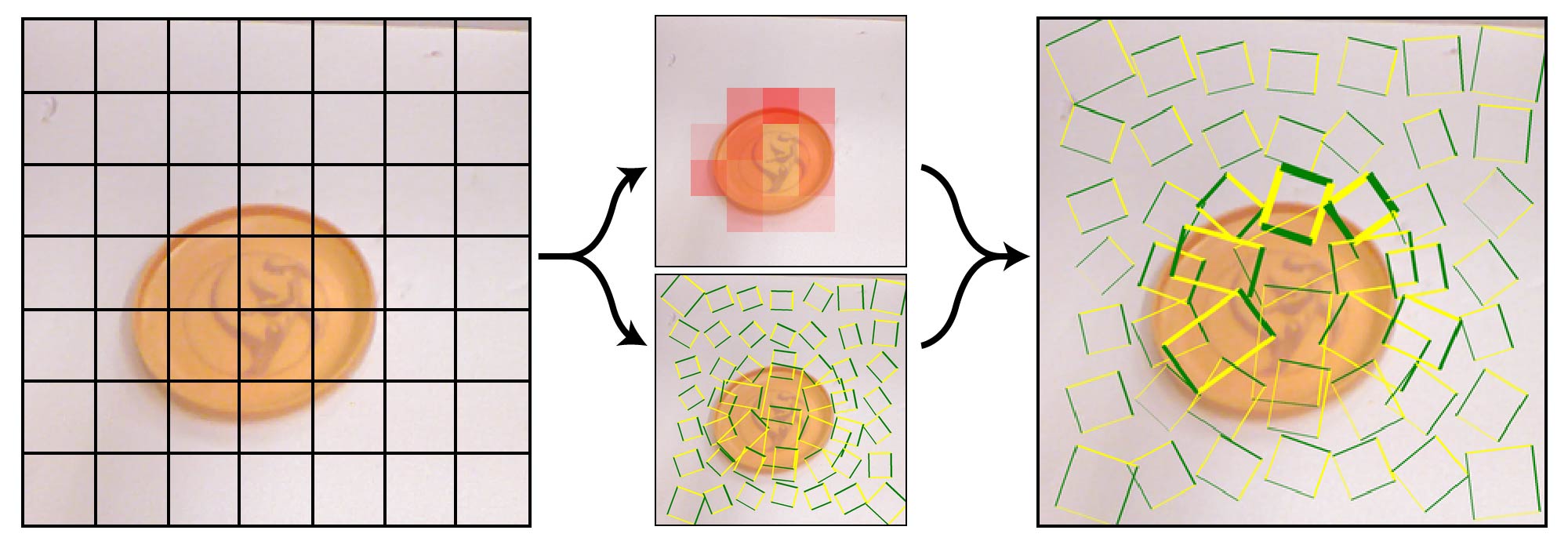}
      \caption{A visualization of the MultiGrasp model running on a test image of a flying disc. The MultiGrasp model splits the image into an NxN grid. For each cell in the grid, the model predicts a bounding box centered at that cell and a probability that this grasp is a true grasp for the object in the image. The predicted bounding boxes are weighted by this probability. The model can predict multiple good grasps for an object, as in this instance. For experiments on the Cornell dataset we pick the bounding box with the highest weight as the final prediction.}
      \label{localized}
   \end{figure*}

\section{EXPERIMENTS AND EVALUATION}

The Cornell Grasping Dataset \cite{cornelldata} contains 885 images of 240 distinct objects and labelled ground truth grasps. Each image has multiple labelled grasps corresponding to different possible ways to grab the object. The dataset is specifically designed for parallel plate grippers. The labels are comprehensive and varied in terms of orientation, location, and scale but they are by no means exhaustive of every possible grasp. Instead they are meant to be diverse examples of particularly good grasps.

\subsection{Grasp Detection}

Previous work uses two different metrics when evaluating grasps on the Cornell dataset. The \textbf{point metric} looks at the distance from the center of the predicted grasp to the center of each of the ground truth grasps. If any of these distances is less than some threshold, the grasp is considered a success.

There are a number of issues with this metric, most notably that it does not consider grasp angle or size. Furthermore, past work does not disclose what values they use for the threshold which makes it impossible to compare new results to old ones. For these reasons we do not evaluate on the point metric.

The second metric considers full grasp rectangles during evaluation. The \textbf{rectangle metric} considers a grasp to be correct if both:
\begin{enumerate}
\item
The grasp angle is within $30^{\circ}$ of the ground truth grasp.
\item 
The Jaccard index of the predicted grasp and the ground truth is greater than 25 percent.
\end{enumerate}
Where the Jaccard index is given by:
\[J(A,B)=\frac{|A\cap B|}{|A\cup B|}\]

The rectangle metric discriminates between good and bad grasps better than the point metric. It is similar to the metrics used in object detection although the threshold on the Jaccard index is lower (25 percent instead of a more standard 50 percent in computer vision) because the ground truth grasps are not exhaustive. A rectangle with the correct orientation that only overlaps by 25 percent with one of the ground truth grasps is still often a good grasp. We perform all of our experiments using the rectangle metric.

Like prior work we use five-fold cross validation for our experimental results. We do two different splits of the data:
\begin{enumerate}
\item \textbf{Image-wise splitting} splits images randomly.
\item \textbf{Object-wise splitting} splits object instances randomly, putting all images of the same object into the same cross-validation split.
\end{enumerate}

Image-wise splitting tests how well the model can generalize to new positions for objects it has seen previously. Object-wise splitting goes further, testing how well the network can generalize to novel objects. In practice, both splitting techniques give comparable performance. This may be due to the similarity between different objects in the dataset (e.g. there are multiple sunglasses of slightly different shapes and colors).

\subsection{Object Classification}

We manually classify the images in the Cornell Grasping Dataset into 16 distinct categories, with categories like ``bottle'', ``shoe'', and ``sporting equipment''. The dataset is not evenly distributed between categories but every category has enough examples in the dataset to be meaningful. The least represented category has 20 images in the dataset while the most represented has 156.

We train and test our combined regression + classification model using these class labels. At test time the combined model simultaneously predicts the best grasp and the object category. We report classification accuracy on the same cross-validation splits as above.

\begin{table*}[tb]
\caption{Rectangle Metric Detection Accuracy on the Cornell Dataset}
\label{results}
\begin{center}
\begin{tabular}{|c|c|c|c|}
\hline
\multirow{2}{*}{\textbf{Algorithm}} & \multicolumn{2}{|c|}{\textbf{Detection accuracy}} & \multirow{2}{*}{\textbf{Time / image}}\\
\hhline{~--~}
& \textbf{Image-wise split} & \textbf{Object-wise split} & \\
\hline
\hline
Chance \cite{lenz2013deep}& 6.7\% & 6.7\% & - \\
\hline
Jiang et al. \cite{lenz2013deep} & 60.5\% & 58.3\% & - \\
\hline
Lenz et al. \cite{lenz2013deep} & 73.9\% & 75.6\% & 13.5 sec\\
\hline
\hline
\textbf{Direct Regression} & 84.4\% & 84.9\% & \multirow{3}{*}{\textbf{76 ms}}\\
\hhline{---~}
\textbf{Regression + Classification} & 85.5\%& 84.9\% & \\
\hhline{---~}
\textbf{MultiGrasp Detection} & \textbf{88.0\%} & \textbf{87.1\%} & \\
\hline
\end{tabular}
\end{center}
\end{table*}

\subsection{Pretraining}

Before training our network on grasps we pretrain on the ImageNet classification task \cite{deng2009imagenet}. Our experience backed by current literature suggests that pretraining large convolutional neural networks greatly improves training time and helps avoid overfitting \cite{oquab2013learning} \cite{donahue2013decaf}.

Krizevsky et al. designed AlexNet for standard RGB images. Low-cost stereo vision systems like the Kinect make RGB-D data increasingly ubiquitous in robotic systems. To use AlexNet with RGB-D data we simply replace the blue channel in the image with the depth information. We could instead modify the architecture to have another input channel but then we would not be able to pretrain the full network.

Pretraining is crucial when there is limited domain-specific data (like labeled RGB-D grasps). Through pretraining the network finds useful, generalizable filters that often translate well to the specific application \cite{donahue2013decaf}. Even in this case where the data format actually changes we still find that the pretrained filters perform well. This may be because good visual filters (like oriented edges) are also good filters in depth space.

\subsection{Training}

We undertake a similar training regimen for each of the models we tested. For each fold of cross-validation, we train each model for 25 epochs. We use a learning rate of 0.0005 across all layers and a weight decay of 0.001. In the hidden layers between fully connected layers we use dropout with a probability of 0.5 as an added form of regularization.

For training and testing our models we use the \texttt{cuda-convnet2} package running on an nVidia Tesla K20 GPU. GPUs offer great benefits in terms of computational power and our timing results depend on using a GPU as part of our pipeline. While GPUs are far from a mainstay in robotic platforms, they are becoming increasingly popular due to their utility in vision tasks.

\subsection{Data Preprocessing}

We perform a minimal amount of preprocessing on the data before feeding it to the network. As discussed previously, the depth information is substituted into the blue channel of the image. The depth information is normalized to fall between 0 and 255. Some pixels lack depth information because they are occluded in the stereo image; we substitute 0 for these pixel values. We then approximately mean-center the image by globally subtracting 144.

When preparing data for training we perform extensive data augmentation by randomly translating and rotating the image. We take a center crop of 320x320 pixels, randomly translate it by up to 50 pixels in both the x and y direction, and rotate it by a random amount. This image is then resized to 224x224 to fit the input layer of our architecture. We generate 3000 training examples per original image. For test images we simply take the center 320x320 crop and resize it without translation or rotation.

\section{RESULTS}

Across the board our models outperform the current state-of-the-art both in terms of accuracy and speed. In Table \ref{results} we compare our results to previous work using their self-reported scores for the rectangle metric accuracy. 

   \begin{figure}[tb]
      \centering
        \includegraphics[width=\linewidth]{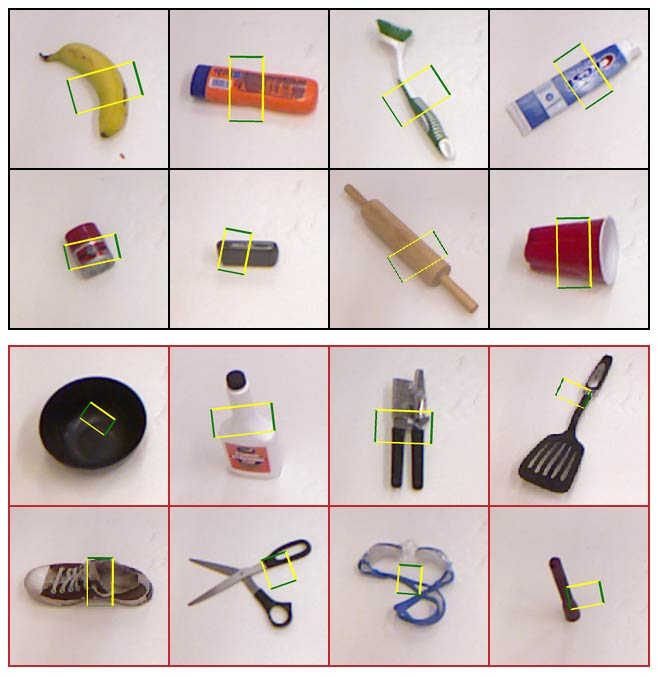}
      \caption{Examples of correct (top) and incorrect (bottom) grasps from the direct regression model. Some incorrect grasps (e.g. the can opener) may actually be viable while others (e.g. the bowl) are clearly not.}
      \label{regression}
   \end{figure}

The direct regression model sets a new baseline for performance in grasp detection. It achieves around 85 percent accuracy in both image-wise and object-wise splits, ten percentage points higher than the previous best. At test time the direct regression model runs in 76 milliseconds per batch, with a batch size of 128 images. While this amounts to processing more than 1,600 images per second, latency matters more than throughput in grasp detection so we report the per batch number as 13 fps. The main source of this speedup is the transition from a scanning window classifier based approach to our single-pass model and our usage of GPU hardware to accelerate computation. 76 milliseconds per frame is certainly achievable on a CPU because it would require only 1/128th of the floating point operations required for processing a full batch on a GPU.

The direct regression model is trained using a different random ground truth grasp every time it sees an image. Due to this it learns to predict the average ground truth grasp for a given object. Predicting average grasps works well with certain types of objects, such as long, thin objects like markers or rolling pins. This model fails mainly in cases where average grasps do not translate to viable grasps on the object, for instance with circular objects like flying discs. Figure \ref{regression} shows some examples of correct and incorrect grasps that the direct regression model predicts.

The combined regression + classification model shows that we can extend our base detection model to simultaneously perform classification without sacrificing detection accuracy; see Table \ref{class} for classification results. Our model can correctly predict the category of an object it has previously seen 9 out of 10 times. When shown novel objects our model predicts the correct category more than 60 percent of the time. By comparison, predicting the most common class would give an accuracy of 17.7 percent.

\begin{table}[htbp]
\caption{Classification Accuracy on the Cornell Dataset}
\label{class}
\begin{center}
\begin{tabular}{|c|c|c|}
\hline
\textbf{Algorithm} & \textbf{Image-wise split} & \textbf{Object-wise split} \\
\hline
\hline
Most Common Class & 17.7\% & 17.7\%\\
\hline
\textbf{Regression + Classification} & 90.0\% & 61.5\% \\
\hline
\end{tabular}
\end{center}
\end{table}

Even with the added classification task the combined model maintains high detection accuracy. It has identical performance on the object-wise split and actually performs slightly better on the image-wise split. This model establishes a strong baseline for combined grasp detection and object classification on the Cornell dataset.

\begin{figure}[tbhp]
      \centering
        \includegraphics[width=\linewidth]{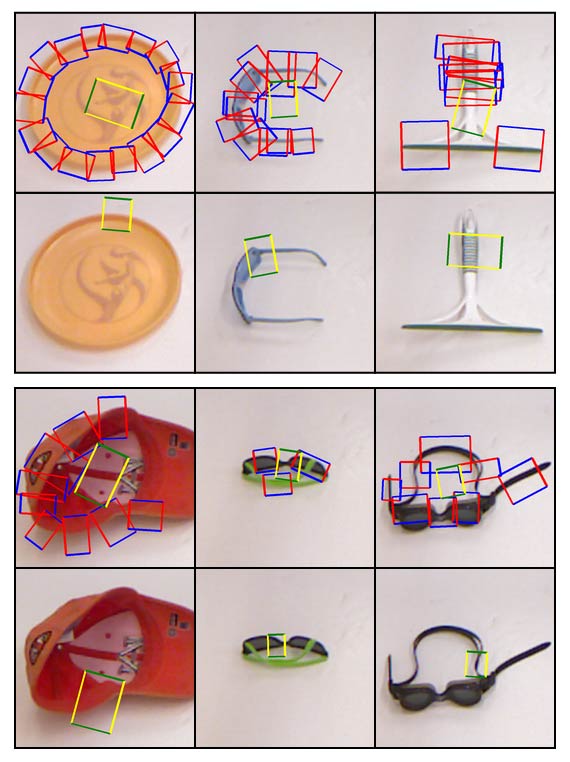}
      \caption{The comparative performance of the direct regression model and MultiGrasp. The top two rows show examples where direct regression model fails due to averaging effects but MultiGrasp predicts a viable grasp. The bottom two rows show examples where both models fail to predict good grasps. The ground truth grasps are shown in blue and red on the direct regression model images.}
      \label{averaging}
   \end{figure}

The MultiGrasp model outperforms our baseline direct regression model by a significant margin. For most objects MultiGrasp gives very similar results to the direct regression model. However, MultiGrasp does not have the same problem with bad average grasps that the direct regression model has which accounts for most of the error reduction. Figure \ref{averaging} shows examples of MultiGrasp outperforming the direct regression model and examples where both models fail.

MultiGrasp has a very similar architecture to the direct regression model and operates at the same real-time speeds. With a grasp detection accuracy of 88 percent and a processing rate of 13 frames per second, MultiGrasp redefines the state-of-the-art in robotic grasp detection.

\section{DISCUSSION}

\addtolength{\textheight}{-3cm}   % This command serves to balance the column lengths

We show that robot perception can be both fast and highly accurate. GPUs provide a large speed boost for visual systems, especially systems based on convolutional neural networks. CNNs continue to dominate other techniques in visual tasks, making GPUs an important component in any high performance robotic system. However, GPUs are most vital during model training and are optimized for throughput, not latency. At test time a CPU could run our model in far less than a second per image, making it viable in real-time robotics applications.

Model consideration is important for achieving high performance. We take advantage of a strong constraint on the data so that our model only needs a single pass over an image to make an accurate grasp prediction.

Our direct regression model uses global information about the image to make its prediction, unlike sliding-window approaches. Sliding window classifiers only see small, local patches thus they can not effectively decide between good grasps and are more easily fooled by false positives. Lenz et al. report very high recognition accuracy for their classifier (94\%) yet it still falls victim to this false positive paradox and its detection accuracy is much lower as a result. In this respect, global models have a large advantage over models that only see local information.

Global models also have their downside. Notably our direct regression model often tries to split the difference between a few good grasps and ends up with a bad grasp. A sliding window approach would never make the mistake of predicting a grasp in the center of a circular object like a flying disc.

Our MultiGrasp model combines the strongest aspects of global and local models. It sees the entire image and can effectively find the best grasp and ignore false positives. However, because each cell can only make a local prediction, it avoids the trap of predicting a bad grasp that falls between several good ones.

The local prediction model also has the ability to predict multiple grasps per image. We are unable to quantitatively evaluate the model in this respect because no current dataset has an appropriate evaluation for multiple grasps in an image. In the future we hope to evaluate this model in a full detection task, either for multiple grasps in an image or on a more standard object detection dataset. 

One further consideration is the importance of pretraining when building large convolutional neural networks. Without pretraining on ImageNet, our models quickly overfit to the training data without learning meaningful representations of good grasps. Interestingly, pretraining worked even across domains and across feature types. We use features tuned for the blue channel of an image on depth information instead and still get good results. Importantly, we get much better results using these features on the depth channel than using them on the original RGB images.

\section{CONCLUSION}

We present a fast, accurate system for predicting robotic grasps of objects in RGB-D images. Our models improve the state-of-the-art and run more than 150 times faster than previous methods. We show that grasp detection and object classification can be combined without sacrificing accuracy or performance. Our MultiGrasp model gets the best known performance on the Cornell Grasping Dataset by combining global information with a local prediction procedure.

\section{ACKNOWLEDGEMENTS}

We would like to thank Alex Krizevsky for helping us with model construction and pretraining, and for helping us customize his \texttt{cuda-convnet2} code. We would also like to thank Vincent Vanhoucke for his insights on model design and for his feedback throughout the experimental process.

\bibliographystyle{IEEEtran}
\bibliography{IEEEabrv,references}

\begin{thebibliography}{10}
\providecommand{\url}[1]{#1}
\csname url@rmstyle\endcsname
\providecommand{\newblock}{\relax}
\providecommand{\bibinfo}[2]{#2}
\providecommand\BIBentrySTDinterwordspacing{\spaceskip=0pt\relax}
\providecommand\BIBentryALTinterwordstretchfactor{4}
\providecommand\BIBentryALTinterwordspacing{\spaceskip=\fontdimen2\font plus
\BIBentryALTinterwordstretchfactor\fontdimen3\font minus
  \fontdimen4\font\relax}
\providecommand\BIBforeignlanguage[2]{{%
\expandafter\ifx\csname l@#1\endcsname\relax
\typeout{** WARNING: IEEEtran.bst: No hyphenation pattern has been}%
\typeout{** loaded for the language `#1'. Using the pattern for}%
\typeout{** the default language instead.}%
\else
\language=\csname l@#1\endcsname
\fi
#2}}

\bibitem{lenz2013deep}
I.~Lenz, H.~Lee, and A.~Saxena, ``Deep learning for detecting robotic grasps,''
  in \emph{Proceedings of Robotics: Science and Systems}, Berlin, Germany, June
  2013.

\bibitem{jiang2011efficient}
Y.~Jiang, S.~Moseson, and A.~Saxena, ``Efficient grasping from rgbd images:
  Learning using a new rectangle representation,'' in \emph{IEEE International
  Conference on Robotics \& Automation (ICRA)}.\hskip 1em plus 0.5em minus
  0.4em\relax IEEE, 2011, pp. 3304--3311.

\bibitem{bicchi2000robotic}
A.~Bicchi and V.~Kumar, ``Robotic grasping and contact: A review,'' in
  \emph{IEEE International Conference on Robotics \& Automation (ICRA)}.\hskip
  1em plus 0.5em minus 0.4em\relax Citeseer, 2000, pp. 348--353.

\bibitem{miller2003automatic}
A.~T. Miller, S.~Knoop, H.~I. Christensen, and P.~K. Allen, ``Automatic grasp
  planning using shape primitives,'' in \emph{IEEE International Conference on
  Robotics \& Automation (ICRA)}, vol.~2.\hskip 1em plus 0.5em minus
  0.4em\relax IEEE, 2003, pp. 1824--1829.

\bibitem{miller2004graspit}
A.~T. Miller and P.~K. Allen, ``Graspit! a versatile simulator for robotic
  grasping,'' \emph{Robotics \& Automation Magazine, IEEE}, vol.~11, no.~4, pp.
  110--122, 2004.

\bibitem{pelossof2004svm}
R.~Pelossof, A.~Miller, P.~Allen, and T.~Jebara, ``An svm learning approach to
  robotic grasping,'' in \emph{IEEE International Conference on Robotics \&
  Automation (ICRA)}, vol.~4.\hskip 1em plus 0.5em minus 0.4em\relax IEEE,
  2004, pp. 3512--3518.

\bibitem{leon2010opengrasp}
B.~Le{\'o}n, S.~Ulbrich, R.~Diankov, G.~Puche, M.~Przybylski, A.~Morales,
  T.~Asfour, S.~Moisio, J.~Bohg, J.~Kuffner, \emph{et~al.}, ``Opengrasp: a
  toolkit for robot grasping simulation,'' in \emph{Simulation, Modeling, and
  Programming for Autonomous Robots}.\hskip 1em plus 0.5em minus 0.4em\relax
  Springer, 2010, pp. 109--120.

\bibitem{lai2011large}
K.~Lai, L.~Bo, X.~Ren, and D.~Fox, ``A large-scale hierarchical multi-view
  rgb-d object dataset,'' in \emph{IEEE International Conference on Robotics \&
  Automation (ICRA)}.\hskip 1em plus 0.5em minus 0.4em\relax IEEE, 2011, pp.
  1817--1824.

\bibitem{lai2012detection}
------, ``Detection-based object labeling in 3d scenes,'' in \emph{IEEE
  International Conference on Robotics \& Automation (ICRA)}.\hskip 1em plus
  0.5em minus 0.4em\relax IEEE, 2012, pp. 1330--1337.

\bibitem{blum2012learned}
M.~Blum, J.~T. Springenberg, J.~Wulfing, and M.~Riedmiller, ``A learned feature
  descriptor for object recognition in rgb-d data,'' in \emph{IEEE
  International Conference on Robotics \& Automation (ICRA)}.\hskip 1em plus
  0.5em minus 0.4em\relax IEEE, 2012, pp. 1298--1303.

\bibitem{henry2010rgb}
P.~Henry, M.~Krainin, E.~Herbst, X.~Ren, and D.~Fox, ``Rgb-d mapping: Using
  depth cameras for dense 3d modeling of indoor environments,'' in \emph{In the
  12th International Symposium on Experimental Robotics (ISER)}.\hskip 1em plus
  0.5em minus 0.4em\relax Citeseer, 2010.

\bibitem{endres2012evaluation}
F.~Endres, J.~Hess, N.~Engelhard, J.~Sturm, D.~Cremers, and W.~Burgard, ``An
  evaluation of the rgb-d slam system,'' in \emph{IEEE International Conference
  on Robotics \& Automation (ICRA)}.\hskip 1em plus 0.5em minus 0.4em\relax
  IEEE, 2012, pp. 1691--1696.

\bibitem{saxena2008robotic}
A.~Saxena, J.~Driemeyer, and A.~Y. Ng, ``Robotic grasping of novel objects
  using vision,'' \emph{The International Journal of Robotics Research},
  vol.~27, no.~2, pp. 157--173, 2008.

\bibitem{rao2010grasping}
D.~Rao, Q.~V. Le, T.~Phoka, M.~Quigley, A.~Sudsang, and A.~Y. Ng, ``Grasping
  novel objects with depth segmentation,'' in \emph{Intelligent Robots and
  Systems (IROS), 2010 IEEE/RSJ International Conference on}.\hskip 1em plus
  0.5em minus 0.4em\relax IEEE, 2010, pp. 2578--2585.

\bibitem{krizhevsky2012imagenet}
A.~Krizhevsky, I.~Sutskever, and G.~E. Hinton, ``Imagenet classification with
  deep convolutional neural networks,'' in \emph{Advances in neural information
  processing systems}, 2012, pp. 1097--1105.

\bibitem{girshick14CVPR}
R.~Girshick, J.~Donahue, T.~Darrell, and J.~Malik, ``Rich feature hierarchies
  for accurate object detection and semantic segmentation,'' in \emph{Computer
  Vision and Pattern Recognition}, 2014.

\bibitem{szegedy2013detection}
\BIBentryALTinterwordspacing
C.~Szegedy, A.~Toshev, and D.~Erhan, ``Deep neural networks for object
  detection,'' in \emph{Advances in Neural Information Processing Systems 26},
  C.~Burges, L.~Bottou, M.~Welling, Z.~Ghahramani, and K.~Weinberger,
  Eds.\hskip 1em plus 0.5em minus 0.4em\relax Curran Associates, Inc., 2013,
  pp. 2553--2561. [Online]. Available:
  \url{http://papers.nips.cc/paper/5207-deep-neural-networks-for-object-detection.pdf}
\BIBentrySTDinterwordspacing

\bibitem{erhan2013scalable}
D.~Erhan, C.~Szegedy, A.~Toshev, and D.~Anguelov, ``Scalable object detection
  using deep neural networks,'' \emph{arXiv preprint arXiv:1312.2249}, 2013.

\bibitem{cornelldata}
``Cornell grasping dataset,''
  \url{http://pr.cs.cornell.edu/grasping/rect_data/data.php}, accessed:
  2013-09-01.

\bibitem{deng2009imagenet}
J.~Deng, W.~Dong, R.~Socher, L.-J. Li, K.~Li, and L.~Fei-Fei, ``Imagenet: A
  large-scale hierarchical image database,'' in \emph{IEEE Conference on
  Computer Vision and Pattern Recognition (CVPR)}.\hskip 1em plus 0.5em minus
  0.4em\relax IEEE, 2009, pp. 248--255.

\bibitem{oquab2013learning}
\BIBentryALTinterwordspacing
M.~Oquab, L.~Bottou, I.~Laptev, and J.~Sivic,
  ``\BIBforeignlanguage{Anglais}{{Learning and Transferring Mid-Level Image
  Representations using Convolutional Neural Networks}},'' in
  \emph{\BIBforeignlanguage{Anglais}{{IEEE Conference on Computer Vision and
  Pattern Recognition}}}, Columbus, OH, {\'E}tats-Unis, Nov. 2013, conference
  version of the paper. [Online]. Available:
  \url{http://hal.inria.fr/hal-00911179}
\BIBentrySTDinterwordspacing

\bibitem{donahue2013decaf}
J.~Donahue, Y.~Jia, O.~Vinyals, J.~Hoffman, N.~Zhang, E.~Tzeng, and T.~Darrell,
  ``Decaf: A deep convolutional activation feature for generic visual
  recognition,'' \emph{arXiv preprint arXiv:1310.1531}, 2013.

\end{thebibliography}

\end{document}